\documentclass[twoside,leqno,twocolumn]{article}
\usepackage{ltexpprt}

\usepackage{graphicx}
\usepackage{natbib}
\usepackage{caption}

\usepackage{algorithm}
\usepackage{algorithmic}
\usepackage{amsfonts}
\usepackage{amsmath}
\usepackage{amssymb}
\usepackage{url}
\usepackage{booktabs}
\usepackage{subcaption}
\usepackage{multirow}
\usepackage[normalem]{ulem}
\usepackage{xcolor}

\title{Adversarial Examples for Extreme Multilabel Text Classification}
\author {
    Mohammadreza Qaraei and
    Rohit Babbar 
}
\date{ Aalto University, Helsinki, Finland}

\def\bbR{\mathbb{R}}
\DeclareMathOperator*{\argmax}{arg\,max}
\def\my{\mathrm{y}}

\begin{document}

\maketitle

\begin{abstract}
Extreme Multilabel Text Classification (XMTC) is a text classification problem in which, (i) the output space is extremely large, (ii) each data point may have multiple positive labels, and (iii) the data follows a strongly imbalanced distribution. 
With applications in recommendation systems and automatic tagging of web-scale documents, the research on XMTC has been focused on improving prediction accuracy and dealing with imbalanced data. However, the robustness of deep learning based XMTC models against adversarial examples has been largely underexplored. \\
In this paper, we investigate the behaviour of XMTC models under adversarial attacks. To this end, first, we define adversarial attacks in multilabel text classification problems. We categorize attacking multilabel text classifiers as (a) positive-targeted, where the target positive label should fall out of top-k predicted labels, and (b) negative-targeted, where the target negative label should be among the top-k predicted labels. Then, by experiments on APLC-XLNet and AttentionXML, we show that XMTC models are highly vulnerable to positive-targeted attacks but more robust to negative-targeted ones.
Furthermore, our experiments show that the success rate of positive-targeted adversarial attacks has an imbalanced distribution. More precisely, tail classes are highly vulnerable to adversarial attacks for which an attacker can generate adversarial samples with high similarity to the actual data-points.
To overcome this problem, we explore the effect of rebalanced loss functions in XMTC where not only do they increase accuracy on tail classes, but they also improve the robustness of these classes against adversarial attacks. The code for our experiments is available at \url{https://github.com/xmc-aalto/adv-xmtc}.
\end{abstract}

\section{Introduction}
\label{sec:intro}

\begin{figure}[t]
\centering
\includegraphics[width=0.47\textwidth]{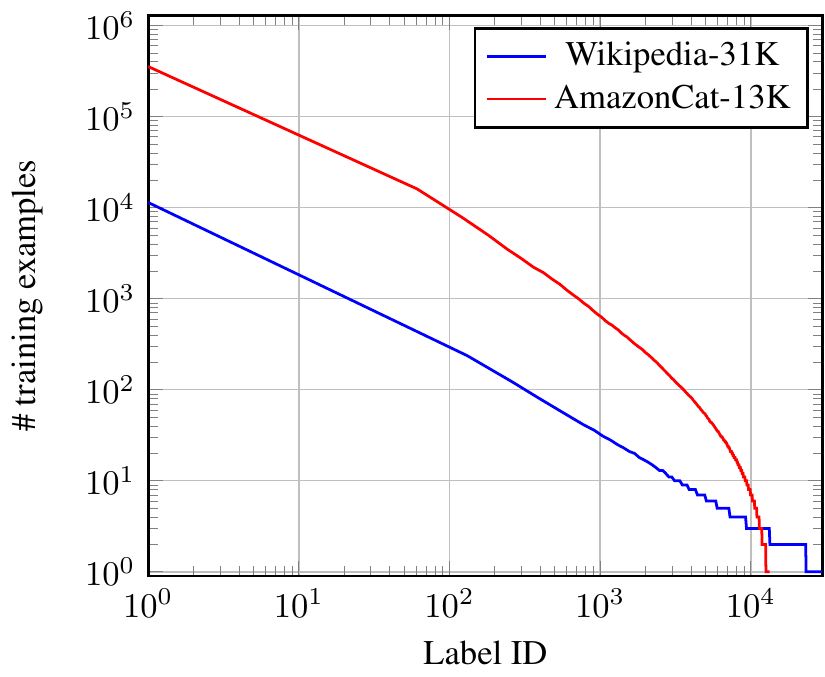}
\caption{
Label frequency of two XMTC datasets, Wikipedia-31K and AmazonCat-13K. Both datasets have an extremely imbalanced distribution, where the frequencies of a few head labels are high, but there are only a few training samples for a large fraction of labels known as tail classes.
}
\label{fig:powerlaw}
\end{figure}

Extreme Multilabel Text Classification (XMTC) addresses the problem of tagging text documents with a few labels from a large label space, which has a wide application in recommendation systems and automatic labelling of web-scale documents \citep{partalas2015lshtc,jain2019slice,agrawal2013multi}. 
There are three characteristics which make XMTC different from typical text classification problems: XMTC is a multilabel problem, the output space is extremely large, and data are highly imbalanced following a power-law distribution \citep{babbar2014power}, which makes models perform poorly on a large fraction of labels with few training samples, known as tail labels (see Figure~\ref{fig:powerlaw}).

The research on XMTC has focused on tackling the aforementioned challenges by proposing models which can scale to millions of labels \citep{babbar2017dismec,jain2019slice,prabhu2018parabel,medini2019extreme} and mitigating the power-law impact on predicting tail classes by rebalancing the loss functions \citep{qaraei2021convex,cui2019class}.
However, as XMTC algorithms have shifted from shallow models on bag-of-words features to deep learning models on word embeddings \citep{you2018attentionxml,ye2020pretrained,jiang2021lightxml}, two new questions need to be addressed : (i) how can one perform adversarial attacks on XMTC models, and (ii) how robust are these models against the generated adversarial examples? 
These questions are also the key to understanding the explainability of modern deep learning models. 

Adversarial attacks are performed by applying engineered noise to a sample, which is imperceptible to humans but can lead deep learning models to misclassify that sample.
While the robustness of deep models to adversarial examples for image classification problems has been extensively studied \citep{szegedy2014intriguing,goodfellow2015explaining}, corresponding methods for generating adversarial examples have also been developed for text classification by taking into account the discrete nature of language data \citep{zhang2020adversarial}. 
However, the research on adversarial attacks on text classifiers is limited to small to medium scale datasets, and the tasks are binary or multiclass problems, making current adversarial frameworks not applicable in XMTC.

In this paper, we explore adversarial attacks on XMTC models. To this end, inspired by \citet{song2018multi} and \citet{hu2021t}, first we define adversarial attacks on multilabel text classification problems. We consider two types of targeted attacks (i) where a sample is manipulated to drop a positive label from the top predicted labels, and (ii) make the model predict a negative label as a positive label. Then we show that XMTC models, in particular the attention-based AttentionXML \citep{you2018attentionxml} and the transformer-based APLC-XLNet \citep{ye2020pretrained}, are vulnerable to adversarial attacks. 

\begin{table}[t]
    \centering
	\begin{tabular}{l l}
		\toprule
		\multirow{6}{*}{\textbf{Sample}} & Gregg shorthand is a form of shorthand \\
		& that was invented by John Robert Gregg \\ & in 1888 ... With the invention of dictation \\
		& machines, shorthand machines, and the \\ & practice of \color{blue}\sout{executives} \color{red}companies \color{black}writing \\ 
		& their own letters on their personal \\ & computers ... \\
		\midrule
		\multirow{4}{*}{\textbf{\begin{tabular}{@{}c@{}}True \\ labels\end{tabular}}} & article, collage, cool, education, graphics, \\ & gregg, language, orthography, reference, \\ & shorthand, speed, stenography, tools, wiki, \\ & wikipedia, writing \\
		\midrule
        \multirow{2}{*}{\textbf{\begin{tabular}{@{}c@{}}Pred. \\ (Real)\end{tabular}}} & language, writing, wikipedia, typography, \\ & \textbf{\color{blue}shorthand\color{black}}, history, wiki, linguistics \\
        \midrule
        \multirow{2}{*}{\textbf{\begin{tabular}{@{}c@{}}Pred. \\ (Adv.)\end{tabular}}} & language, writing, wikipedia, typography, \\ &history, wiki, \textbf{\color{red}shorthand\color{black}}, english \\
		\bottomrule
	\end{tabular}
	\caption{An adversarial example generated for APLC-XLNet by targeting the tail label ``shorthand" of the Wikipedia-31K dataset. While ``shorthand" is among top-5 predicted labels for the real sample, it will become the 7th predicted label by only replacing ``executives'' with ``companies". Notably, the newly predicted label ``history" is not one of the true labels.}
\label{tbl:adv-example-intro}
\end{table}

Our analysis also shows that the success rate of the adversarial attacks on XMTC models has imbalanced behaviour, similar to the distribution of the data.
In particular, our experiments show that positive tail classes are very easy to attack. This means that not only is it difficult to correctly predict a tail label, but also there is a high chance that one can eliminate a correctly classified tail label from the predicted labels by changing a few words, or even a single word in some cases (Table~\ref{tbl:adv-example-intro}). 

To improve robustness of tail classes against adversarial attacks, we investigate the rebalanced loss functions originally proposed to enhance model performance on missing/infrequent labels \citep{qaraei2021convex}.
Our results show these loss functions can significantly increase robustness of correctly (when trained with vanilla loss) classified samples belonging to tail classes.

To summarize, the key findings of our work are:

\begin{itemize}
    \item XMTC models are vulnerable to adversarial examples, as shown by experiments with AttentionXML and APLC-XLNet.
    \item The success rate of the adversarial attacks on XMTC has an imbalanced behaviour similar to the distribution of data, where it is easy to attack positive tail labels by only changing a few words in the corresponding samples.
    \item The rebalanced loss functions can significantly improve the robustness of tail labels against adversarial attacks.
\end{itemize}

\section{Related work}
\label{sec:related-work}

\subsection{Adversarial attacks on text classifiers}
Adversarial attacks on image classification cannot be directly applied to text classification problems because of the discrete structure of text data. In text classification problems, this is achieved by first finding ``important" parts of the text and then manipulating these parts.
In white-box attacks, the important parts are determined by the gradient information, and in black-box attacks this is done by masking some parts of the text and then computing the difference between the output probabilities of the masked and unmasked sample.
A perturbation that is undetectable for humans should result in a sample that is semantically similar to the original and preserve the fluency of the text.

Adversarial attacks in text classification problems can be categorized into character-level and word-level attacks.
In \citet{sun2020adv} and \citet{li2018textbugger}, which are two character-level attacks, first, important words are determined by the gradient magnitude and then those words are misspelled to change the label.
The same idea is used in \citet{gao2018black} but with a black-box setting for finding the important words. The problem with the character-level methods is that a spell checker can easily reveal the adversarial samples. 

In word-level attacks, \citet{jin2020bert} and \citet{ren2019generating} find the important words in a black-box setting and then replace those words with synonyms to change the predicted label. 
However, the substituted words may damage the fluency of the sentences.
To preserve the fluency of the sentences, a language model such as BERT \citep{devlin2018bert} can be used to generate the candidates in a context-aware setting \citep{garg2020bae,li2020bert,xu2021attacking}.

Current works in adversarial attacks on text classifiers have focused on binary or multiclass problems without taking into account data irregularities such as imbalanced data distribution for instance.
For attacking XMTC models, first we extend the adversarial framework for finding important words to multilabel settings, then we use the BERT model \citep{li2020bert} for context-aware word substitutions which can generate fluent adversarial samples.

\subsection{XMTC models}

Earlier works in XMTC used shallow models on bag-of-words features \citep{bhatia2015sparse, babbar2017dismec, khandagale2019bonsai}. 
However, as bag-of-words representation looses contextual information, recent XMTC models employ deep neural networks on word embbedings. 
Among these models, AttentionXML \citep{you2018attentionxml} uses a BiLSTM layer followed by an attention module over pretrained word embbedings and is trained in a tree-structure to reduce the computational complexity. 
APLC-XLNet \citep{ye2020pretrained} is a transformer-based approach, which fine tunes XLNet \citep{yang2019xlnet} on extreme classification datasets. To scale XLNet to a large number of labels, APLC-XLNet partitions labels based on their frequencies, and the loss for most of the samples is computed only on a fraction of these partitions.

Another major challenge in XMTC is the problem of infrequent and missing labels \citep{Jain16,qaraei2021convex}. 
To improve generalization on infrequent labels, ProXML \citep{babbar2019data} optimizes squared hinge loss with $\ell_1$ regularization. 
To address the missing labels problem, \citet{Jain16} optimizes a propensity scored variant of normalized discounted cumulative gain (nDCG) in a tree classifier. 
\citet{qaraei2021convex} propose to reweight popular loss functions, such as BCE and squared hinge loss to make them convex surrogates for the unbiased 0-1 loss. The reweighted loss functions are further rebalanced by a function of label frequencies to improve performance on tail classes. 
Our experiments show that, even though these losses were not designed from a robustness perspective but more from the viewpoint of being statistically unbiased under missing labels, they significantly improve the robustness of tail classes against adversarial attacks in deep XMTC models.

\subsection{Adversarial attacks on imbalanced or multilabel problems}

Adversarial attacks on multilabel problems were first defined in \citet{song2018multi} for multilabel classification or ranking. \citet{hu2021t} used a different approach for attacking multilabel models by proposing loss functions which are based on the top-k predictions. In \citet{melacci2020can}, the domain knowledge on the relationships among different classes are used to evade adversarial attacks against multilabel problems. \citet{yang2020characterizing} defined the attackability of mulitlabel classifiers, and proved that the spectral norm of a classifier's parameters and its performance on unperturbed data are two key factors in this regard.

Adversarial attacks on models trained on imbalanced data were discussed in \citet{wang2021imbalanced} and \citet{wu2021adversarial}. In both works, it has been remarked that for adversarially trained models, the decay of accuracy from head to tail classes on both clean and adversarial examples are more than that of normal training. To overcome this problem, \citet{wu2021adversarial} used a margin-based scale-invariant loss to deal with imbalanced distribution, along with a loss to control the robustness of the model. \citet{wang2021imbalanced} showed that rebalancing robust training can increase the accuracy of tail classes but has significant adverse effects on head classes. To tackle this problem, they proposed to use reweighted adversarial training along with a loss which makes features more separable.

All the aforementioned works are on image classification problems. 
To the best of our knowledge, adversarial attacks on multilabel problems in the text classification domain have only been explored in \citet{wu2017adversarial}, which did not aim to produce adversarial examples, but to use adversarial training to improve accuracy. 

\section{Adversarial attacks on multilabel text classifiers}
\label{sec:adv-multilabel}

Attacking multilabel problems is different from binary or multiclass problems since the samples in the first may have multiple positive labels, and therefore the corresponding manipulated sample can be an adversarial sample for some labels but not for the others. While attacking multilabel image classification models has been recently explored in \citet{song2018multi} and \citet{hu2021t}, to the best of our knowledge, there is no work on attacking multilabel problems in the NLP domain. In this section, we formally define attacking multilabel text classifiers.

Assume $S=[w_1, ..., w_n]$ is a document consisting of $n$ words $w_i \in \bbR^d$, $\my \in \{0,1\}^L$ are the labels corresponding to this document in one-hot encoded format, and $Y=\{i\rvert y_i=1\}$ represents indices of the positive labels. Let $g: \bbR^{d\times n} \rightarrow \bbR^L$ be a mapping from documents to scores, where $g_i(S)\in R$ indicates the score of the $i$-th label. Also, $\hat{Y}_k(S)=\{T_1(g(s)),...,T_k(g(s))\}$ represents the top-k predicted labels, where $T_i:\bbR^L\rightarrow \{1,...,L\}$ is an operator which returns the index of the $i$-th largest value. The goal in adversarial attacks is to generate a document $S^\prime$ which is similar to $S$ but has different predicted labels\footnote{In our setting, we measure similarity of $S$ and $S^\prime$ by the number of different words in them, and also the similarity of their word embbedings by Universal Sentence Encoder \citep{cer2018universal}}.

As in multiclass classification, we can also have non-targeted and targeted attacks on multilabel problems, which are defined in the following.

\subsection*{Non-targeted attacks:}
In a non-targeted attack, the goal is to remove at least one correctly predicted positive label from the top-k predictions;
A modified document $S^\prime$ is an adversarial example for $S$ if $S^\prime$ is similar to $S$ and $\exists i \in \hat{Y}_k(S)$ such that $i \in Y$ and $i \notin \hat{Y}_k(S^\prime)$.

\subsection*{Targeted attacks:}
In a targeted attack on a multilabel problem, an attacker may try to decrease the scores of some particular positive labels, or increase the scores of some negative labels in order to be among the top-k predicted labels. To this end, we use a subset of positive labels denoted by $A_1 \subset \{i \rvert y_i=1\}$ and a subset of negative labels $A_{-1} \subset \{i \rvert y_i=0\}$ to define which labels are targeted. This leads to two types of attacks:
\begin{itemize}
    \item Positive-targeted attacks:
    
        $\forall i \in A_1 \cap \hat{Y}_k(S): i \notin \hat{Y}_K(S^\prime)$
        
    \item Negative-targeted attacks:
    
        $\forall i \in A_{-1}\setminus \hat{Y}_k(S): i \in \hat{Y}_k(S^\prime)$

\end{itemize}

In general, attacking text classification models consists of two steps: first, finding important words, and second, manipulating the most important words in order to change the labels.
In the subsequent paragraphs, following \citet{jin2020bert} for a black-box attack using Bert, we describe how to find important words in a sequence based on the target of the attack, and how to perform word substitution.

\subsection{Finding important words}
In a black-box attack, the only information available from the model is the output scores. We compute the importance of each word by masking that word in the document and measuring how much we are closer to our goal based on the target of the attack and the output scores.

Formally, assume $S=[w_1, ..., w_n]$ is a document, and $S_{\setminus w_i}=[w_1, ...,w_{i-1}, \text{[MASK]} ,w_{i+1}, ...]$ is the document in which the $i$-th word is masked. In a non-targeted attack, the importance of the $i$-th word is computed as follows:

\begin{equation}
    \label{eq:important-non-target}
    I_{w_i} = \sum_{l \in \hat{Y}_k(S)} g_{l}(S) - g_{l}(S_{\setminus w_i})
\end{equation}
This equation assigns an importance score to word $w_i$ by summing the drop in the scores of predicted labels when that word is masked.

Similarly, in positive targeted attacks, the important words should be those which decrease the output scores of correctly predicted labels in $A_1$ more than other words when they are masked. Hence, the importance of the $i$-th word is computed as follows:

\begin{equation}
    \label{eq:important-pos-target}
    I^p_{w_i} = \sum_{a_1 \in A_1 \cap \hat{Y}_k(S)} g_{a_1}(S) - g_{a_1}(S_{\setminus w_i})
\end{equation}

Furthermore, for negative-targeted attacks, the importance of the $i$-th word is computed as the sum of the difference of the output scores of the labels in $A_{-1}$, if they are not among the predicted labels, after masking the $i$-th word:

\begin{equation}
    \label{eq:important-neg-target}
    I^n_{w_i} = \sum_{a_{-1} \in A_{-1} \setminus \hat{Y}_k(S)} g_{a_{-1}}(S_{\setminus w_i}) - g_{a_{-1}}(S)
\end{equation}

\subsection{Word substitution}

Since word substitution can be the same for multiclass and multilabel problems, we can use the existing methods to replace important words. We use Bert model \citep{devlin2018bert,li2020bert} for this purpose which leads to a context-aware method and produces fluent adversarial samples. To this end, we mask the important words of a sample one by one and pass that sample to a Bert model to generate candidates for the masked words. In each trial $t$, we pick the word suggested by the Bert model for which the difference between the output scores towards our goal is maximized. For non-targeted and positive-targeted attacks, this is obtained by:

\begin{equation}
    \label{eq:pick-word}
    w^t = \argmax_k \sum_{j\in \Gamma} g_j(S^{t-1}) - g_j(S_{w^t_k}^{t-1})
\end{equation}
where $S^{t-1}$ is the sample after changing $t-1$ important words, and $S^{t-1}_{w^t_k}$ is that sample when the $t$-th important word is replaced by the $k$-th suggested word from Bert.
Also, $\Gamma$ is $\hat{Y}_k(S)$ in a non-targeted attack, or equal to $A_1 \cap \hat{Y}_k(S)$ in a positive-targeted attack.
Moreover, for negative-targeted attacks, $w^t$ is computed as Equation~\ref{eq:pick-word}, but the sum should be multiplied by a negative sign and $\Gamma = A_{-1} \setminus \hat{Y}_k(S)$.

We repeat masking the important words and feeding them to the network until the goal for the attack is reached, or we are out of the limit of the allowed number of changes. A pseudocode for positive-targeted attacks is given in Algorithm~\ref{alg:cap}.

\begin{algorithm}[t]
\caption{Positive-targeted attack pseudocode}\label{alg:cap}
\textbf{Input:}  target label set $A_1$, sample $S=[w_1, ..., w_n]$, score function $g(.)$, label predictor function $\hat{Y}_k(.)$, a Bert language model, maximum allowed change rate $\theta$\\
\textbf{Output:} Adversarial sample $S^\prime$ \\
\begin{algorithmic}[1]
\STATE $\Gamma \gets A_1 \cap \hat{Y}_k(S)$
\FOR{$i=1,...,n$} 
    \STATE $S_{\setminus w_i} \gets [w_1, ...,w_{i-1}, \text{[MASK]}, ...,w_n]$
    \STATE $I^p_{w_i} \gets \sum_{a_1 \in \Gamma} g_{a_1}(S) - g_{a_1}(S_{\setminus w_i})$
\ENDFOR
\STATE $I=argsort([I^p_{w_1}, ..., I^p_{w_n}])$ \COMMENT{sorting in descending order}
\STATE $t \gets 1$, $S^0 \gets S$
\WHILE{$t<\theta$}
    \STATE $i \gets I[t]$
    \STATE $H \gets$ suggested words by Bert for $S^{t-1}$ in which the $i$-th word is masked
    \STATE $d \gets 0$
    \FOR{$w_k$ in $H$}
        \STATE $S^{t-1}_{w_k^t} \gets [w_1^{t-1}, ..., w_{i-1}^{t-1}, \{w_k\}, ..., w_n^{t-1}]$
        \STATE $d_0 \gets \sum_{j\in \Gamma} g_j(S^{t-1}) - g_j(S_{w^t_k}^{t-1})$
        \IF{$d_0>d$}
            \STATE $w^t \gets w_k$, $d \gets d_0$
        \ENDIF
    \ENDFOR
    \STATE $S^t \gets [w_1^{t-1}, ..., w_{i-1}^{t-1}, \{w^t\}, ..., w_n^{t-1}]$
    \STATE  $S^\prime \gets S^t$, $t \gets t+1$
    \IF{$\hat{Y}_k(S^\prime) \cap A_1 = \emptyset$}
        \STATE return $S^\prime$
    \ENDIF
\ENDWHILE
\STATE \textbf{return} None
\end{algorithmic}
\end{algorithm}

\section{Adversarial attacks on XMTC models}
\label{sec:adv-xmc}

In this section, firstly we show that XMTC models are vulnerable to positive-targeted but more robust to negative-targeted attacks. An important observation about positive-targeted attacks is that their success rate has an imbalanced distribution, where one can successfully attack a tail label by changing only a few words in the document, while head classes are more robust to the attacks.

Secondly, to increase the robustness of tail classes against adversarial attacks, we replace the normal loss functions with the rebalanced variants \citep{qaraei2021convex} in the targeted models. The results show that these loss functions can significantly improve the robustness of tail classes.

\subsection{Setup}
\label{sec:setup}

Adversarial attacks are performed on two XMTC models, AttentionXML and APLC-XLNet, trained on two extreme classification datasets, AmazonCat-13K and Wikipedia-31K \citep{Bhatia16}. The statistics of these datasets are shown in Table~\ref{tbl:datasets}. Similar to other datasets in XMTC, both datasets follow an extremely imbalanced distribution (Figure~\ref{fig:powerlaw}).

\begin{table*}[t]
    \centering
	\begin{tabular}{l|r|r|r|r|r}
		\toprule
		\textbf{Dataset} &  \# \textbf{Training } &  \# \textbf{Test } & \# \textbf{Labels} & \textbf{APpL} & \textbf{ALpP} \\ \midrule
\textbf{AmazonCat-13K} &  1,186,239  & 306,782  & 13,330 & 448.5 & 5.04\\
\textbf{Wikipedia-31K} &  14,146  & 6,616  & 30,938 & 8.5 & 18.6\\
		\bottomrule
	\end{tabular}
	\caption{The statistics of AmazonCat-13K and Wikipedia-31K \cite{Bhatia16}. APpL and ALpP denote the average points per label and the average labels per point, respectively.}
\label{tbl:datasets}
\end{table*}

\begin{table*}[t]
    \centering
	\begin{tabular}{l|c|c|c|c}
		\toprule
		\textbf{Model} & \textbf{Dataset} & \textbf{\begin{tabular}{@{}c@{}}Success \\ rate (\%)\end{tabular}} & \textbf{Similarity} & \textbf{\begin{tabular}{@{}c@{}}Change \\ rate (\%)\end{tabular}} \\ 
		\midrule
\multirow{2}{*}{\textbf{APLC-XLNet}} & \textbf{Wikipedia-31K} & 89.94 & 0.74 & 1.14 \\
& \textbf{AmazonCat-13K} & 85.12 & 0.68 & 1.95 \\
        \midrule
        \midrule
\multirow{2}{*}{\textbf{AttentionXML}} & \textbf{Wikipedia-31K} & 96.66 & 0.83 & 0.35 \\
& \textbf{AmazonCat-13K} & 84.40 & 0.67 & 2.20 \\
		\bottomrule
	\end{tabular}
	\caption{The success rate of positive-targeted adversarial attacks against APLC-XLNet and AttentionXML on Wikipedia-31K and AmazonCat-13K. The success rate is more than 80\% for all the cases with a high similarity between the adversarial and real samples and small change rate.}
\label{tbl:positive-results}
\end{table*}

We only perform positive-targeted or negative-targeted attacks, since these types of attacks are more practical in real-world problems than non-targeted attacks \citep{song2018multi}, and give us the opportunity to compare the behaviour of the models under attacking classes with different frequencies. 

We only consider the case where the target set contains a single label. Also, for each target label, we consider the samples in which the target label is among (not among) the true positive labels in positive-targeted (negative-targeted) attacks.
We randomly draw the samples in which the target label is classified correctly. 
It means that the accuracy of the models on the drawn samples with respect to the target labels is always perfect in both of the attacks.

To treat labels with different frequencies equally, we partition label frequencies in different bins and draw an equal number of samples for each bin for all the experiments.
We consider several consecutive frequencies as one bin, if there are at least $L$ labels in that bin for which there is at least one correctly classified sample for each of the labels, where $L=100$ for Wikipedia-31K and is $400$ for AmazonCat-13K.

To measure how much the adversarial samples are similar to the original samples, we use two criteria: i) cosine similarity of the encoded samples using Universal Sentence Encoder (USE) \citep{cer2018universal} which gives us a measure in $[0,1]$, ii) change rate, which is the percentage of the words changed in a real sample to generate an adversarial sample.

\subsection{General results}
\label{sec:aplc-results}

\subsubsection{Positive-targeted attacks}
The results of positive-targeted adversarial attacks on APLC-XLNet and AttentionXML for about $1000$ samples uniformly drawn from different label frequency bins are shown in Table~\ref{tbl:positive-results}. Here the maximum allowed change rate is set to 10\%. As the results indicate, 
the success rate of the positive-targeted attacks against both models is high, which is more than 90\% for Wikipedia-31K and more than 84\% for AmazonCat-13K. Furthermore, the generated samples are similar to the real samples in terms of USE similarity, and the change rate is less than 2.5\% in all the cases.

Overall, the experiments show that XMTC models are vulnerable to positive-targeted attacks where an adversary can fool the model not to predict a particular label by a few changes in the document.

\subsubsection{Negative-targeted attacks}

While positive-targeted attacks have high success rates on both models, having a high success rate for negative-targeted attacks is not easy. This is due to the fact that finding the words which can increase the prediction probability of a particular label from the extremely large vocabulary space and injecting them into a document is much harder than finding the words inside a document that can lead to a lower probability for a label and replacing them with semantically similar words (positive-targeted attacks).

\begin{table*}[t]
    \centering
	\begin{tabular}{c|c}
		\toprule
		\textbf{Dataset} & \textbf{Clusters}\\
		\midrule
        \multirow{4}{*}{\textbf{AmazonCat-13K}} & \begin{tabular}{@{}c@{}} authorship, book industry, editing, play \& scriptwriting, \\ newspapers \& magazines, children's literature \end{tabular}\\
        \cmidrule{2-2}
        & \begin{tabular}{@{}c@{}} sailing, water sports, ships, canoeing, solo travel, \\ transportation, canada \end{tabular}\\
        \midrule
        \multirow{2}{*}{\textbf{Wikipedia-31K}} & t-test, significance, cs483, correlation\\
        \cmidrule{2-2}
        & cmd, variable, batchfiles, command \\
		\bottomrule
	\end{tabular}
	\caption{Four clusters of the AmazonCat-13K and Wikipedia-31K labels. In our negative-targeted attacks, a sample may be a candidate to attack, if it has at least one positive label of those which are in the same cluster as the target label.}
\label{tbl:cluster-amazoncat}
\end{table*}

\begin{table*}[t]
    \centering
	\begin{tabular}{l|c|c|c|c|c}
		\toprule
		\textbf{Model} & \textbf{Dataset} &  \textbf{Clustering} & \textbf{\begin{tabular}{@{}c@{}}Success \\ rate (\%)\end{tabular}} & \textbf{Similarity} & \textbf{\begin{tabular}{@{}c@{}}Change \\ rate (\%)\end{tabular}} \\ 
		\midrule
\multirow{4}{*}{\textbf{APLC-XLNet}} & \multirow{2}{*}{\textbf{W}} & N & 1.67 & 0.00  & 7.35 \\
& & Y & 16.29 & 0.11  & 5.27 \\
\cmidrule{2-6}
& \multirow{2}{*}{\textbf{AC}} & N & 49.09 & 0.20 & 3.77 \\
& & Y & 53.70 & 0.30 & 3.39\\
        \midrule
        \midrule
\multirow{4}{*}{\textbf{AttentionXML}} & \multirow{2}{*}{\textbf{W}} & N & 36.53 & 0.15 & 3.31 \\
& & Y & 38.00 & 0.12 & 3.41 \\
\cmidrule{2-6}
& \multirow{2}{*}{\textbf{AC}} & N & 1.95 & 0.00 & 4.39 \\
& & Y & 12.06 & 0.08 & 5.20\\
        
		\bottomrule
	\end{tabular}
	\caption{The success rate of negative-targeted adversarial attacks against APLC-XLNet and AttentionXML on Wikipedia-31K (\textbf{W}) and AmazonCat-13K (\textbf{AC}). While the both models show robustness to the adversarial attacks, when the samples to attack are restricted to those which have at least one positive label in the same cluster as the target label, the success rate of the attack is significantly higher than the naive case.}
\label{tbl:negative-results}
\end{table*}

To have higher success rates for negative-targeted attacks, for each target label, we restrict the attack to the samples for which the label is close to those samples but they don't contain that label as a positive label. In our work, we assume that a label is close to a sample if that sample has at least one positive label which is in the same cluster as the target label. We perform the clustering by the balanced hierarchical binary clustering of \citet{prabhu2018parabel}, where each label is represented by the sum of the TF-IDF representation of the documents for which that label is a positive label. Formally, assume $S_1, ..., S_N$ are our documents in the training set and $X=[x_1, ..., x_N]^T\in \bbR^{N\times V}$ are the corresponding TF-IDF representations of these documents. Also, $Z \in \{0,1\}^{N \times L}$ consists of the one-hot labels for each document. Then $\hat{z}_l = z_l \times X$ is the representation that we use for the $l$-th label to perform clustering, where $z_l$ is the $l$-th row of $Z$. Some of the clusters for AmazonCat-13K are depicted in Table~\ref{tbl:cluster-amazoncat}.

After the clustering is done, for each target label $l \in C_k$ where $C_k$ is the $k$-th cluster, we consider only the following samples to attack:
\begin{equation}
    \label{eq:cluster-attack}
    NT_l = \{S_i \rvert i\in\{1,...,N\}, Y(S_i)\cap {C_k}_{\setminus l} \neq \varnothing \}
\end{equation}
where $Y(S_i)$ consists of the indices of positive labels for the document $S_i$.

For negative-targeted attacks, the number of random samples that we draw for each bin is different among different datasets and models, and it is equal to the minimum number of samples among the bins which meet the conditions\footnote{For the negative-targeted attacks, the conditions for picking a sample for a particular target label are: (i) the target label should not be among the positive labels of that sample, (ii) the target label should not be in the top-k predicted labels, and (iii) the sample should have at least one label inside the same cluster as the target label}. Also, the cluster size is set to at least 3 labels. The results of the negative-targeted attacks are presented in Table~\ref{tbl:negative-results}. Here the results are with and without using clustering for drawing the samples for each target label. As the results show, the two XMTC models are robust to negative-targeted attacks. However, the difference in the success rate of using and not using clustering for picking the samples to attack is significant, where it can be more than 10\% in some cases. 

We should note that clustering of labels has been used in many XMTC works but for increasing the speed of training and evaluating the model \citep{prabhu2018parabel,khandagale2019bonsai,jiang2021lightxml,Mittal2021DECAFDE}.

\subsection{Label-frequency-based results}
In this subsection, first, we analyse how the success rate of the positive-targeted attacks changes with respect to data distribution. Second, we investigate the effect of using rebalanced loss functions on this trend.

\subsubsection{Attacking labels with different frequencies}

\begin{table*}[t]
    \centering
	\begin{tabular}{c|c|l|c|c|c}
		\toprule
		\textbf{Model} & \textbf{Dataset} & \textbf{Sample} & \textbf{\begin{tabular}{@{}c@{}}Target \\ Label\end{tabular}} & \textbf{RR} & \textbf{RA}\\
		\midrule
        \multirow{10}{*}{\rotatebox[origin=c]{90}{\textbf{APLC-XLNet}}} & \textbf{W} & \begin{tabular}{@{}l@{}l@{}} Security-Enhanced Linux (SELinux) \\ is a Linux \color{blue}\sout{feature} \color{red}system \color{black}that provides \\ a variety of security policies, including \\ U.S. Department of Defense style \\ mandatory access controls ...\end{tabular} & selinux & 5 & 7\\
        \cmidrule{2-6}
        & \textbf{AC} & \begin{tabular}{@{}l@{}l@{}} This studded protective vinyl chair \\ \color{blue}\sout{mat} \color{red}Mat \color{black}is top of the line ... \end{tabular} & chair mats & 5 & 6\\
        \cmidrule{2-6}
        & \textbf{AC} & \begin{tabular}{@{}l@{}l@{}}Mar-Hyde One-Step Rust Converter \\ Primer Sealer chemically reacts to ... \\ Also proven effective on new degreased \\ or mild \color{blue}\sout{steel} \color{red}iron \color{black}where flash-rusting \\ has occurred ...\end{tabular} & \begin{tabular}{@{}c@{}}corrosion \& \\ rust \\ inhibitors \end{tabular} & 4 & 6\\
        
        \midrule
        \multirow{10}{*}{\rotatebox[origin=c]{90}{\textbf{AttentionXML}}} & \textbf{W} & \begin{tabular}{@{}l@{}l@{}} the following are lists of notable people \\ who intentionally terminated their own \\ lives ... the following is a list of people \\ whose cause of \color{blue}\sout{death} \color{red}deaths \color{black}is disputed \\ or whose intention to commit suicide is \\ doubtful ...\end{tabular} & morte & 4 & 6\\
        \cmidrule{2-6}
        & \textbf{AC} & \begin{tabular}{@{}l@{}l@{}} leviton pr180-1lw decora 500w \\ incandescent 400va passive infrared \\ wall switch occupancy \color{blue}\sout{sensor} \\ \color{red}camera \color{black} single pole and 3-way white ... \end{tabular} & \begin{tabular}{@{}c@{}} motion-\\activated \\ switches \end{tabular} & 5 & 6\\
        \cmidrule{2-6}
        & \textbf{AC} & \begin{tabular}{@{}l@{}l@{}}30 built in trim kitrim for r530es- \\ r530bs stainle ... with the \color{blue}\sout{sharp} \\ \color{red}acute \color{black} rk51s30 stainless 30-inch \\ trim kit ...\end{tabular} & \begin{tabular}{@{}c@{}}trim kits \end{tabular} & 4 & 6\\

		\bottomrule
	\end{tabular}
	\caption{Several adversarial samples targeted tail classes from Wikipedia-31K (\textbf{W}) and AmazonCat-13K (\textbf{AC}).
    \textbf{RR} stands for the rank of targeted labels for the real samples, and \textbf{RA} is the rank of those labels for the adversarial samples. In all the cases, the targeted tail labels fall out of top-5 predicted labels by changing only one word.}
\label{tbl:adv-example-exp}
\end{table*}

\begin{figure*}[!ht]
\centering
\begin{subfigure}{.5\textwidth}
  \centering
  \includegraphics[width=.9\linewidth]{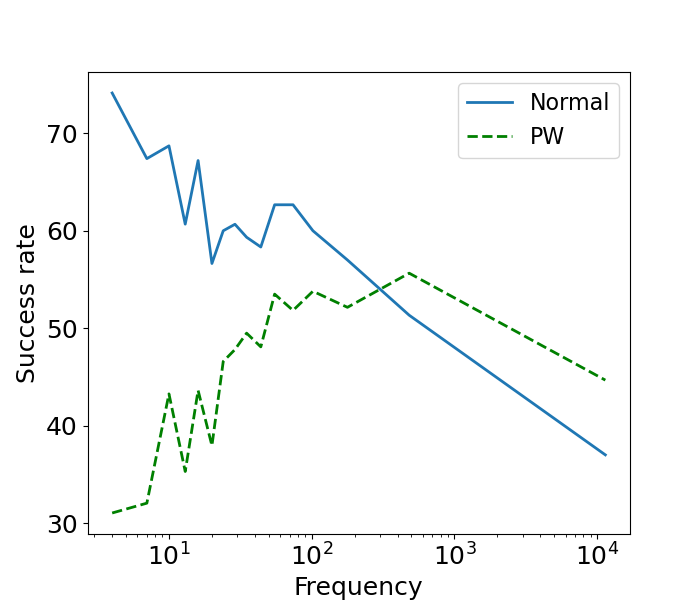}
  \caption{APLC-XLNet (Wikipedia-31K)}
  \label{fig:attack-wiki-aplc-un}
\end{subfigure}%
\begin{subfigure}{.5\textwidth}
  \centering
  \includegraphics[width=.9\linewidth]{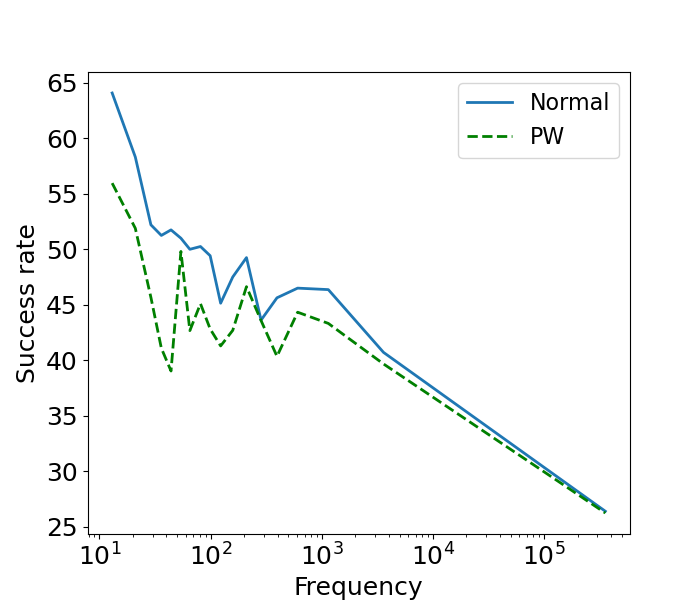}
  \caption{APLC-XLNet (AmazonCat-13K)}
  \label{fig:attack-amazon-aplc-un}
\end{subfigure}
\quad
\begin{subfigure}{.5\textwidth}
  \centering
  \includegraphics[width=.9\linewidth]{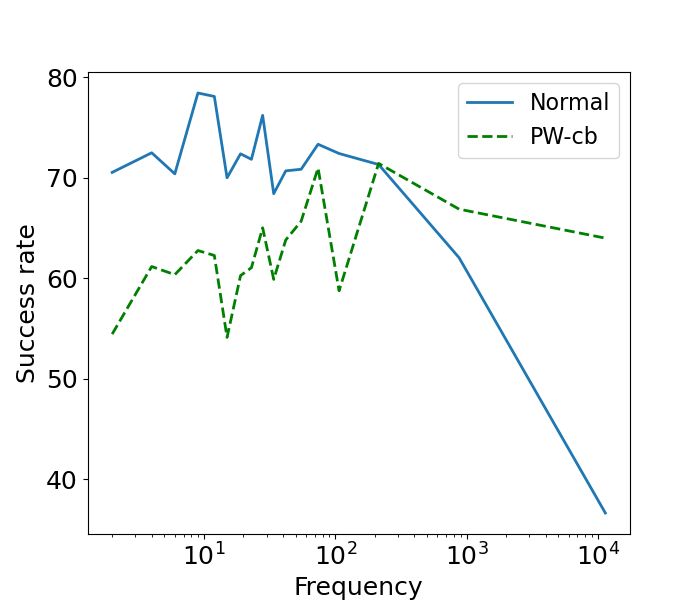}
  \caption{AttentionXML (Wikipedia-31K)}
  \label{fig:attack-wiki-axml-un}
\end{subfigure}%
\begin{subfigure}{.5\textwidth}
  \centering
  \includegraphics[width=.9\linewidth]{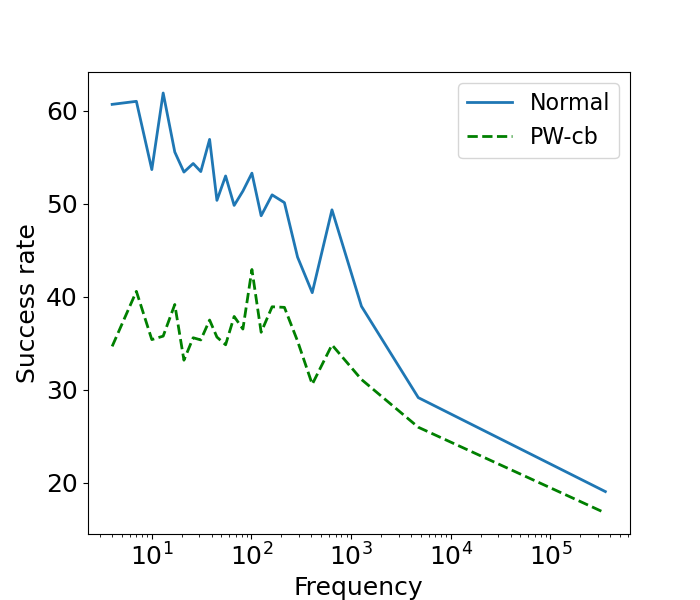}
  \caption{AttentionXML (AmazonCat-13K)}
  \label{fig:attack-amazon-axml-un}
\end{subfigure}
\caption{The success rate of positive-targeted attacks against APLC-XLNet and AttentionXML trained on two XMTC datasets for different label frequencies. An attack is successful if the similarity of the real and adversarial samples is above 0.8 and the change rate is lower than 10\%. For the normal loss functions, the success rate exhibits an imbalanced behaviour, where the higher values are for the lower frequencies. Rebalanced loss functions mitigate this problem by improving the robustness of tail classes against these attacks.}
\label{fig:attack-unbiased}
\end{figure*}

The success rate of positive-targeted adversarial attacks on labels with different frequencies are demonstrated in Figure~\ref{fig:attack-unbiased} (graphs labeled with ``Normal''). We follow the setup introduced in the Setup section to categorize labels in different bins based on their frequencies, and the number of randomly drawn samples for each bin is set to 200 for Wikipedia-31K and to 600 for AmazonCat-13K. Also, for these experiments, an attack is considered successful if the USE similarity of the generated adversarial samples with the real samples is above $0.8$, and the change rate is less than $10\%$. It means that the generated adversarial samples are highly similar to the corresponding real samples. 

As the figures show, the success rate of the attacks on both datasets and models has an imbalanced behaviour, where the gap between the tail and head classes is more than $30\%$ in all the cases. It shows that it is easy to generate an adversarial sample for a tail label with a high similarity to the real samples, while this practice becomes difficult for the head classes. Some generated samples for tail classes are depicted in Table~\ref{tbl:adv-example-exp}. 

We would like to remind the reader that in our experiments, all the samples used for generating the adversarial attacks are classified correctly. It implies the fact that besides the challenge of predicting tail labels correctly, these labels are also more vulnerable to adversarial attacks when they are correctly predicted.

\subsubsection{Robust XMTC with unbiased-rebalanced losses}

The rebalanced loss functions are originally proposed for the problem of missing labels and imbalanced data \citep{qaraei2021convex}. These loss functions, for losses which decompose over labels, such as the hinge and BCE loss, suggest the following form:

\begin{equation}
    \label{eq:unbiased-loss}
    l(\my,\hat{\my}) = \sum_{j=1}^{L} C_j W_j l^{+}(y_j,\hat{y}_j) + l^{-}(y_j,\hat{y}_j)
\end{equation}
where $l^+$ ($l^-$) is the positive (negative) part of the original loss. Also, $C_j$ is a factor to rebalance the loss, and $W_j$ is a factor to compensate for missing labels.

Suggested by \citet{qaraei2021convex}, we set $C_j = \frac{1-\beta}{1-\beta^{n_j}}$ \citep{cui2019class} where $\beta=0.9$, and $W_j = 2/p_j-1$ where $p_j$, called the propensity score of label $j$, indicates the probability of the label being present and is computed by the empirical model of \citet{Jain16}. While $C_j$ is explicitly introduced to rebalance the loss, $W_j$ also reweights the loss in favor of tail classes as the problem of missing labels is more pervasive in those classes.

Figure~\ref{fig:attack-unbiased} demonstrates a comparison of the original models with the rebalanced variants under the adversarial attacks. Here we refer to the loss modified for missing labels (only using $W_j$) as PW, and when the rebalancing factor ($C_j$) is also taken into account, we call the method PW-cb. In our experiments with rebalanced loss functions, the choice of the type of reweighting for each dataset and model depends on its prediction performance.

To compare the normal loss with the reweighted variants under the adversarial attacks, we attack the samples that are classified correctly by the normal loss, but when a reweighted loss is used to train the model.

As the figures show, the rebalanced variants significantly improve the robustness of the models on less frequent classes. This means that using the reweighted loss functions improves the robustness of the model on the samples that are classified correctly by the normal loss but missclassified after performing the attack. The gap is large for all the model and datasets, between $10\%$ to $40\%$, for the least frequent classes.

We should remark that although the reweighted loss functions improve robustness of tail classes against adversarial attacks, they have an adverse effect on head classes in Wikipedia-31K dataset. This is mostly due to two labels ``wiki'' and ``wikipedia'' which exist in more than $87\%$ and $81\%$ of samples, and have tiny weights in the reweighted loss functions because of having very high frequencies.

\section{Conclusion}
\label{sec:conclusion}

In this paper, we investigated adversarial attacks on extreme multilabel text classification (XMTC) problems. Due to the multilabel setting and extremely imbalanced data in these problems, the settings and responses for the adversarial attacks are different from the typical text classification problems.
We observed that XMTC models are vulnerable to adversarial attacks when an attacker tries to remove a specific true label of a sample from the set of predicted labels, which are called positive-targeted attacks. Also, our findings show that, besides the difficulty of correctly predicting tail classes, a new challenge in XMTC that should be considered is the low robustness of these classes against adversarial attacks. We showed that this problem can be mitigated by using the unbiased-rebalanced loss functions which reweight the loss in favour of tail classes. Some remaining questions include whether there are ways to efficiently attack XMTC models by targeting negative labels, and also how to adversarially train an XMTC model given that generating adversarial examples, which needs multiple running of the Bert model for each sample, and adding them to the clean data causes tremendous additional costs to the computationally expensive training of these models.

\bibliographystyle{plainnat}

\begin{thebibliography}{39}
\providecommand{\natexlab}[1]{#1}
\providecommand{\url}[1]{\texttt{#1}}
\expandafter\ifx\csname urlstyle\endcsname\relax
  \providecommand{\doi}[1]{doi: #1}\else
  \providecommand{\doi}{doi: \begingroup \urlstyle{rm}\Url}\fi

\bibitem[Agrawal et~al.(2013)Agrawal, Gupta, Prabhu, and
  Varma]{agrawal2013multi}
Rahul Agrawal, Archit Gupta, Yashoteja Prabhu, and Manik Varma.
\newblock Multi-label learning with millions of labels: Recommending advertiser
  bid phrases for web pages.
\newblock In \emph{Proceedings of the 22nd international conference on World
  Wide Web}, pages 13--24, 2013.

\bibitem[Babbar and Sch{\"o}lkopf(2017)]{babbar2017dismec}
Rohit Babbar and Bernhard Sch{\"o}lkopf.
\newblock Dismec: Distributed sparse machines for extreme multi-label
  classification.
\newblock In \emph{Proceedings of the Tenth ACM International Conference on Web
  Search and Data Mining}, pages 721--729, 2017.

\bibitem[Babbar and Sch{\"o}lkopf(2019)]{babbar2019data}
Rohit Babbar and Bernhard Sch{\"o}lkopf.
\newblock Data scarcity, robustness and extreme multi-label classification.
\newblock \emph{Machine Learning}, 108\penalty0 (8):\penalty0 1329--1351, 2019.

\bibitem[Babbar et~al.(2014)Babbar, Metzig, Partalas, Gaussier, and
  Amini]{babbar2014power}
Rohit Babbar, Cornelia Metzig, Ioannis Partalas, Eric Gaussier, and Massih-Reza
  Amini.
\newblock On power law distributions in large-scale taxonomies.
\newblock \emph{ACM SIGKDD explorations newsletter}, 16\penalty0 (1):\penalty0
  47--56, 2014.

\bibitem[Bhatia et~al.(2016)Bhatia, Dahiya, Jain, Kar, Mittal, Prabhu, and
  Varma]{Bhatia16}
K.~Bhatia, K.~Dahiya, H.~Jain, P.~Kar, A.~Mittal, Y.~Prabhu, and M.~Varma.
\newblock The extreme classification repository: Multi-label datasets and code,
  2016.

\bibitem[Bhatia et~al.(2015)Bhatia, Jain, Kar, Varma, and
  Jain]{bhatia2015sparse}
Kush Bhatia, Himanshu Jain, Purushottam Kar, Manik Varma, and Prateek Jain.
\newblock Sparse local embeddings for extreme multi-label classification.
\newblock In \emph{NIPS}, 2015.

\bibitem[Cer et~al.(2018)Cer, Yang, Kong, Hua, Limtiaco, John, Constant,
  Guajardo-C{\'e}spedes, Yuan, Tar, et~al.]{cer2018universal}
Daniel Cer, Yinfei Yang, Sheng-yi Kong, Nan Hua, Nicole Limtiaco, Rhomni~St
  John, Noah Constant, Mario Guajardo-C{\'e}spedes, Steve Yuan, Chris Tar,
  et~al.
\newblock Universal sentence encoder.
\newblock \emph{arXiv preprint arXiv:1803.11175}, 2018.

\bibitem[Cui et~al.(2019)Cui, Jia, Lin, Song, and Belongie]{cui2019class}
Yin Cui, Menglin Jia, Tsung-Yi Lin, Yang Song, and Serge Belongie.
\newblock Class-balanced loss based on effective number of samples.
\newblock In \emph{Proceedings of the IEEE Conference on Computer Vision and
  Pattern Recognition}, pages 9268--9277, 2019.

\bibitem[Devlin et~al.(2018)Devlin, Chang, Lee, and Toutanova]{devlin2018bert}
Jacob Devlin, Ming-Wei Chang, Kenton Lee, and Kristina Toutanova.
\newblock Bert: Pre-training of deep bidirectional transformers for language
  understanding.
\newblock \emph{arXiv preprint arXiv:1810.04805}, 2018.

\bibitem[Gao et~al.(2018)Gao, Lanchantin, Soffa, and Qi]{gao2018black}
Ji~Gao, Jack Lanchantin, Mary~Lou Soffa, and Yanjun Qi.
\newblock Black-box generation of adversarial text sequences to evade deep
  learning classifiers.
\newblock In \emph{2018 IEEE Security and Privacy Workshops (SPW)}, pages
  50--56. IEEE, 2018.

\bibitem[Garg and Ramakrishnan(2020)]{garg2020bae}
Siddhant Garg and Goutham Ramakrishnan.
\newblock Bae: Bert-based adversarial examples for text classification.
\newblock In \emph{Proceedings of the 2020 Conference on Empirical Methods in
  Natural Language Processing (EMNLP)}, pages 6174--6181, 2020.

\bibitem[Goodfellow et~al.(2015)Goodfellow, Shlens, and
  Szegedy]{goodfellow2015explaining}
Ian Goodfellow, Jonathon Shlens, and Christian Szegedy.
\newblock Explaining and harnessing adversarial examples.
\newblock In \emph{International Conference on Learning Representations}, 2015.

\bibitem[Hu et~al.(2021)Hu, Ke, Wang, and Lyu]{hu2021t}
Shu Hu, Lipeng Ke, Xin Wang, and Siwei Lyu.
\newblock T$_k${ML-AP}: Adversarial attacks to top-k multi-label learning.
\newblock \emph{arXiv preprint arXiv:2108.00146}, 2021.

\bibitem[Jain et~al.(2016)Jain, Prabhu, and Varma]{Jain16}
Himanshu Jain, Yashoteja Prabhu, and Manik Varma.
\newblock Extreme multi-label loss functions for recommendation, tagging,
  ranking and other missing label applications.
\newblock In \emph{KDD}, August 2016.

\bibitem[Jain et~al.(2019)Jain, Balasubramanian, Chunduri, and
  Varma]{jain2019slice}
Himanshu Jain, Venkatesh Balasubramanian, Bhanu Chunduri, and Manik Varma.
\newblock Slice: Scalable linear extreme classifiers trained on 100 million
  labels for related searches.
\newblock In \emph{Proceedings of the Twelfth ACM International Conference on
  Web Search and Data Mining}, pages 528--536, 2019.

\bibitem[Jiang et~al.(2021)Jiang, Wang, Sun, Yang, Zhao, and
  Zhuang]{jiang2021lightxml}
Ting Jiang, Deqing Wang, Leilei Sun, Huayi Yang, Zhengyang Zhao, and Fuzhen
  Zhuang.
\newblock Lightxml: Transformer with dynamic negative sampling for
  high-performance extreme multi-label text classification.
\newblock In \emph{Proceedings of the AAAI Conference on Artificial
  Intelligence}, volume~35, pages 7987--7994, 2021.

\bibitem[Jin et~al.(2020)Jin, Jin, Zhou, and Szolovits]{jin2020bert}
Di~Jin, Zhijing Jin, Joey~Tianyi Zhou, and Peter Szolovits.
\newblock Is bert really robust? a strong baseline for natural language attack
  on text classification and entailment.
\newblock In \emph{Proceedings of the AAAI conference on artificial
  intelligence}, volume~34, pages 8018--8025, 2020.

\bibitem[Khandagale et~al.(2020)Khandagale, Xiao, and
  Babbar]{khandagale2019bonsai}
Sujay Khandagale, Han Xiao, and Rohit Babbar.
\newblock Bonsai: diverse and shallow trees for extreme multi-label
  classification.
\newblock \emph{Machine Learning}, pages 1--21, 2020.

\bibitem[Li et~al.(2018)Li, Ji, Du, Li, and Wang]{li2018textbugger}
Jinfeng Li, Shouling Ji, Tianyu Du, Bo~Li, and Ting Wang.
\newblock Textbugger: Generating adversarial text against real-world
  applications.
\newblock \emph{arXiv preprint arXiv:1812.05271}, 2018.

\bibitem[Li et~al.(2020)Li, Ma, Guo, Xue, and Qiu]{li2020bert}
Linyang Li, Ruotian Ma, Qipeng Guo, Xiangyang Xue, and Xipeng Qiu.
\newblock Bert-attack: Adversarial attack against bert using bert.
\newblock In \emph{Proceedings of the 2020 Conference on Empirical Methods in
  Natural Language Processing (EMNLP)}, pages 6193--6202, 2020.

\bibitem[Medini et~al.(2019)Medini, Huang, Wang, Mohan, and
  Shrivastava]{medini2019extreme}
Tharun Kumar~Reddy Medini, Qixuan Huang, Yiqiu Wang, Vijai Mohan, and Anshumali
  Shrivastava.
\newblock Extreme classification in log memory using count-min sketch: A case
  study of amazon search with 50m products.
\newblock \emph{Advances in Neural Information Processing Systems},
  32:\penalty0 13265--13275, 2019.

\bibitem[Melacci et~al.(2020)Melacci, Ciravegna, Sotgiu, Demontis, Biggio,
  Gori, and Roli]{melacci2020can}
Stefano Melacci, Gabriele Ciravegna, Angelo Sotgiu, Ambra Demontis, Battista
  Biggio, Marco Gori, and Fabio Roli.
\newblock Can domain knowledge alleviate adversarial attacks in multi-label
  classifiers?
\newblock \emph{arXiv preprint arXiv:2006.03833}, 2020.

\bibitem[Mittal et~al.(2021)Mittal, Dahiya, Agrawal, Saini, Agarwal, Kar, and
  Varma]{Mittal2021DECAFDE}
Anshul Mittal, Kunal Dahiya, Sheshansh Agrawal, Deepak Saini, Sumeet Agarwal,
  Purushottam Kar, and Manik Varma.
\newblock Decaf: Deep extreme classification with label features.
\newblock \emph{Proceedings of the 14th ACM International Conference on Web
  Search and Data Mining}, 2021.

\bibitem[Partalas et~al.(2015)Partalas, Kosmopoulos, Baskiotis, Artieres,
  Paliouras, Gaussier, Androutsopoulos, Amini, and Galinari]{partalas2015lshtc}
Ioannis Partalas, Aris Kosmopoulos, Nicolas Baskiotis, Thierry Artieres, George
  Paliouras, Eric Gaussier, Ion Androutsopoulos, Massih-Reza Amini, and Patrick
  Galinari.
\newblock Lshtc: A benchmark for large-scale text classification.
\newblock \emph{arXiv preprint arXiv:1503.08581}, 2015.

\bibitem[Prabhu et~al.(2018)Prabhu, Kag, Harsola, Agrawal, and
  Varma]{prabhu2018parabel}
Yashoteja Prabhu, Anil Kag, Shrutendra Harsola, Rahul Agrawal, and Manik Varma.
\newblock Parabel: Partitioned label trees for extreme classification with
  application to dynamic search advertising.
\newblock In \emph{Proceedings of the 2018 World Wide Web Conference}, pages
  993--1002, 2018.

\bibitem[Qaraei et~al.(2021)Qaraei, Schultheis, Gupta, and
  Babbar]{qaraei2021convex}
Mohammadreza Qaraei, Erik Schultheis, Priyanshu Gupta, and Rohit Babbar.
\newblock Convex surrogates for unbiased loss functions in extreme
  classification with missing labels.
\newblock In \emph{Proceedings of the Web Conference 2021}, pages 3711--3720,
  2021.

\bibitem[Ren et~al.(2019)Ren, Deng, He, and Che]{ren2019generating}
Shuhuai Ren, Yihe Deng, Kun He, and Wanxiang Che.
\newblock Generating natural language adversarial examples through probability
  weighted word saliency.
\newblock In \emph{Proceedings of the 57th annual meeting of the association
  for computational linguistics}, pages 1085--1097, 2019.

\bibitem[Song et~al.(2018)Song, Jin, Huang, and Hu]{song2018multi}
Qingquan Song, Haifeng Jin, Xiao Huang, and Xia Hu.
\newblock Multi-label adversarial perturbations.
\newblock In \emph{2018 IEEE International Conference on Data Mining (ICDM)},
  pages 1242--1247. IEEE, 2018.

\bibitem[Sun et~al.(2020)Sun, Hashimoto, Yin, Asai, Li, Yu, and
  Xiong]{sun2020adv}
Lichao Sun, Kazuma Hashimoto, Wenpeng Yin, Akari Asai, Jia Li, Philip Yu, and
  Caiming Xiong.
\newblock Adv-bert: Bert is not robust on misspellings! generating nature
  adversarial samples on bert.
\newblock \emph{arXiv preprint arXiv:2003.04985}, 2020.

\bibitem[Szegedy et~al.(2014)Szegedy, Zaremba, Sutskever, Bruna, Erhan,
  Goodfellow, and Fergus]{szegedy2014intriguing}
Christian Szegedy, Wojciech Zaremba, Ilya Sutskever, Joan Bruna, Dumitru Erhan,
  Ian Goodfellow, and Rob Fergus.
\newblock Intriguing properties of neural networks.
\newblock In \emph{International Conference on Learning Representations}, 2014.

\bibitem[Wang et~al.(2021)Wang, Xu, Liu, Li, Thuraisingham, and
  Tang]{wang2021imbalanced}
Wentao Wang, Han Xu, Xiaorui Liu, Yaxin Li, Bhavani Thuraisingham, and Jiliang
  Tang.
\newblock Imbalanced adversarial training with reweighting.
\newblock \emph{arXiv preprint arXiv:2107.13639}, 2021.

\bibitem[Wu et~al.(2021)Wu, Liu, Huang, Wang, and Lin]{wu2021adversarial}
Tong Wu, Ziwei Liu, Qingqiu Huang, Yu~Wang, and Dahua Lin.
\newblock Adversarial robustness under long-tailed distribution.
\newblock In \emph{Proceedings of the IEEE/CVF Conference on Computer Vision
  and Pattern Recognition}, pages 8659--8668, 2021.

\bibitem[Wu et~al.(2017)Wu, Bamman, and Russell]{wu2017adversarial}
Yi~Wu, David Bamman, and Stuart Russell.
\newblock Adversarial training for relation extraction.
\newblock In \emph{Proceedings of the 2017 Conference on Empirical Methods in
  Natural Language Processing}, pages 1778--1783, 2017.

\bibitem[Xu and Veeramachaneni(2021)]{xu2021attacking}
Lei Xu and Kalyan Veeramachaneni.
\newblock Attacking text classifiers via sentence rewriting sampler.
\newblock \emph{arXiv preprint arXiv:2104.08453}, 2021.

\bibitem[Yang et~al.(2019)Yang, Dai, Yang, Carbonell, Salakhutdinov, and
  Le]{yang2019xlnet}
Zhilin Yang, Zihang Dai, Yiming Yang, Jaime Carbonell, Ruslan Salakhutdinov,
  and Quoc~V Le.
\newblock Xlnet: Generalized autoregressive pretraining for language
  understanding.
\newblock \emph{arXiv preprint arXiv:1906.08237}, 2019.

\bibitem[Yang et~al.(2020)Yang, Han, and Zhang]{yang2020characterizing}
Zhuo Yang, Yufei Han, and Xiangliang Zhang.
\newblock Characterizing the evasion attackability of multi-label classifiers.
\newblock \emph{arXiv preprint arXiv:2012.09427}, 2020.

\bibitem[Ye et~al.(2020)Ye, Chen, Wang, and Davison]{ye2020pretrained}
Hui Ye, Zhiyu Chen, Da-Han Wang, and Brian Davison.
\newblock Pretrained generalized autoregressive model with adaptive
  probabilistic label clusters for extreme multi-label text classification.
\newblock In \emph{International Conference on Machine Learning}, pages
  10809--10819. PMLR, 2020.

\bibitem[You et~al.(2019)You, Zhang, Wang, Dai, Mamitsuka, and
  Zhu]{you2018attentionxml}
Ronghui You, Zihan Zhang, Ziye Wang, Suyang Dai, Hiroshi Mamitsuka, and
  Shanfeng Zhu.
\newblock Attentionxml: Label tree-based attention-aware deep model for
  high-performance extreme multi-label text classification.
\newblock In \emph{NeurIPS}, pages 5812--5822, 2019.

\bibitem[Zhang et~al.(2020)Zhang, Sheng, Alhazmi, and Li]{zhang2020adversarial}
Wei~Emma Zhang, Quan~Z Sheng, Ahoud Alhazmi, and Chenliang Li.
\newblock Adversarial attacks on deep-learning models in natural language
  processing: A survey.
\newblock \emph{ACM Transactions on Intelligent Systems and Technology (TIST)},
  11\penalty0 (3):\penalty0 1--41, 2020.

\end{thebibliography}


\end{document}